\documentclass[10pt,twocolumn,letterpaper]{article}

\usepackage[pagenumbers]{cvpr} % To force page numbers, e.g. for an arXiv version

\usepackage{graphicx}
\usepackage{amsmath}
\usepackage{amssymb}
\usepackage{booktabs}
\usepackage{color}
\usepackage{colortbl}
\usepackage[accsupp]{axessibility}  % Improves PDF readability for those with disabilities.

\usepackage[pagebackref,breaklinks,colorlinks]{hyperref}
\usepackage{cvpr}      % To produce the REVIEW version
\usepackage{graphicx}
\usepackage{amsmath}
\usepackage{amssymb}
\usepackage{booktabs}
\usepackage{color}
\usepackage{multicol}
\usepackage{multirow}
\usepackage{array}

\usepackage[capitalize]{cleveref}
\crefname{section}{Sec.}{Secs.}
\Crefname{section}{Section}{Sections}
\Crefname{table}{Table}{Tables}
\crefname{table}{Tab.}{Tabs.}

\def\cvprPaperID{1587} % *** Enter the CVPR Paper ID here
\def\confName{CVPR}
\def\confYear{2023}
\definecolor{color3}{rgb}{0.95,0.95,0.95}

\begin{document}

%%%%%%%%% TITLE - PLEASE UPDATE
\title{Large-capacity and Flexible Video Steganography via Invertible Neural Network}

% \author{First Author\\
% Institution1\\
% Institution1 address\\
% {\tt\small firstauthor@i1.org}
% % For a paper whose authors are all at the same institution,
% % omit the following lines up until the closing ``}''.
% % Additional authors and addresses can be added with ``\and'',
% % just like the second author.
% % To save space, use either the email address or home page, not both
% \and
% Second Author\\
% Institution2\\
% First line of institution2 address\\
% {\tt\small secondauthor@i2.org}
% }
\author{
Chong Mou$^{1}$, Youmin Xu$^{1}$, Jiechong Song$^{1}$, Chen Zhao$^{2}$, Bernard Ghanem$^{2}$, Jian Zhang$^{1, 3, *}$\\
$^1$Peking University Shenzhen Graduate School, Shenzhen, China\\
$^2$King Abdullah University of Science and Technology (KAUST), Thuwal, Saudi Arabia\\
$^3$Peng Cheng Laboratory, Shenzhen, China\\
% {\tt\small eechongm@gmail.com; lucywq1028@gmail.com; zhangjian.sz@pku.edu.cn}
% eechongm@stu.pku.edu.cn; zhangjian.sz@pku.edu.cn
\tt\small \{eechongm, youmin.xu\}@stu.pku.edu.cn; \{songjiechong, zhangjian.sz\}@pku.edu.cn;\\
\tt\small \{chen.zhao, Bernard.Ghanem\}@kaust.edu.sa
}
\maketitle
\let\thefootnote\relax\footnotetext{$^*$\textit{Corresponding author}. This work was supported by the King Abdullah University of Science and Technology (KAUST) Office of Sponsored Research through the Visual Computing Center (VCC) funding, SDAIA-KAUST Center of Excellence in Data Science and Artificial Intelligence, and Shenzhen Research Project JCYJ20220531093215035.}

%%%%%%%%% ABSTRACT
\begin{abstract}
Video steganography is the art of unobtrusively concealing secret data in a cover video and then recovering the secret data through a decoding protocol at the receiver end. Although several attempts have been made, most of them are limited to low-capacity and fixed steganography. To rectify these weaknesses, we propose a \textbf{L}arge-capacity and \textbf{F}lexible \textbf{V}ideo \textbf{S}teganography \textbf{N}etwork (LF-VSN) in this paper. For large-capacity, we present a reversible pipeline to perform multiple videos hiding and recovering through a single invertible neural network (INN). Our method can \textbf{hide/recover 7 secret videos in/from 1 cover video} with promising performance. For flexibility, we propose a key-controllable scheme, enabling different receivers to recover particular secret videos from the same cover video through specific keys. Moreover, we further improve the flexibility by proposing a scalable strategy in multiple videos hiding, which can hide variable numbers of secret videos in a cover video with a single model and a single training session. Extensive experiments demonstrate that with the significant improvement of the video steganography performance, our proposed LF-VSN has high security, large hiding capacity, and flexibility. The source code is available at \url{https://github.com/MC-E/LF-VSN}.
% \vspace{-6pt}
\end{abstract}

\begin{figure*}[t]
% \small
    \centering
    \includegraphics[width=0.98\linewidth]{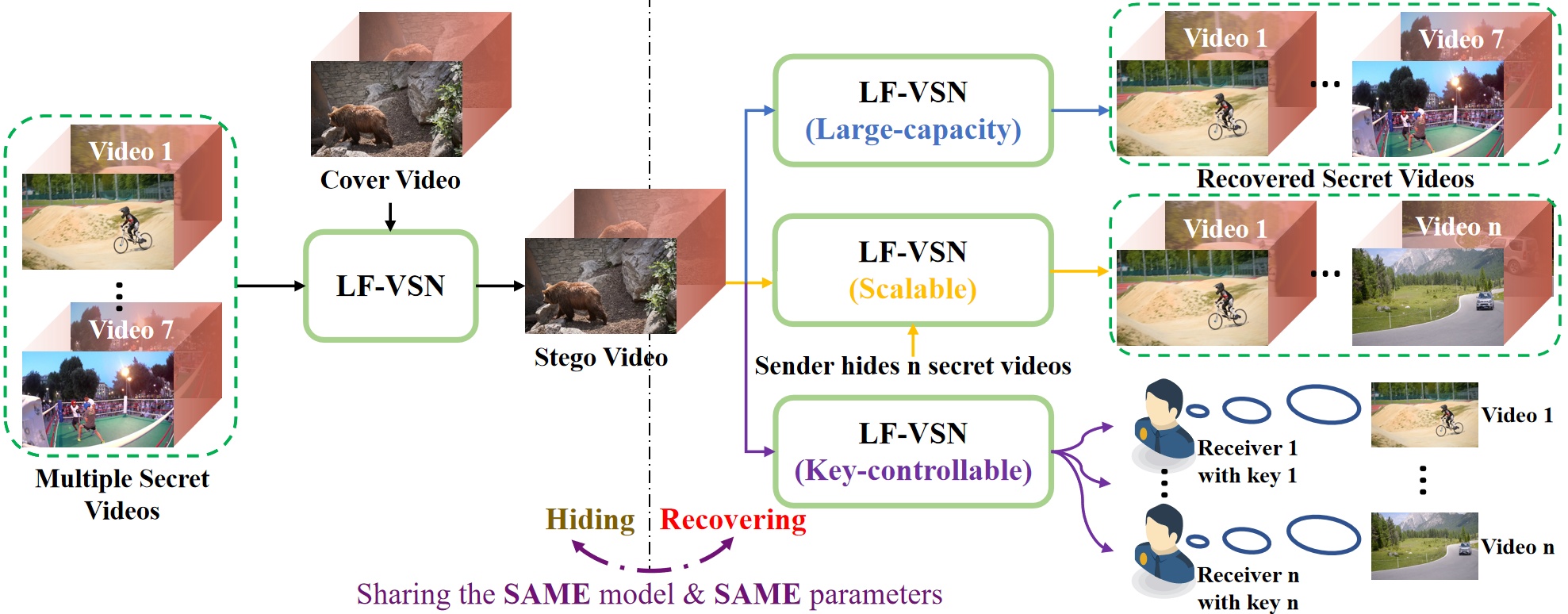}
    % \vspace{4pt}
    \caption{Illustration of our large-capacity and flexible video steganography network (LF-VSN). Our LF-VSN reversibly solves multiple videos hiding and recovering with a single model and the same parameters. It has large-capacity, key-controllable and scalable advantages.
    }
    \vspace{-10pt}
    \label{overview}
\end{figure*}
\section{Introduction}
Steganography~\cite{survey} is the technology of hiding some secret data into an inconspicuous cover medium to generate a stego output, which only allows the authorized receiver to recover the secret information. Unauthorized people can only access the content of the plain cover medium, and hard to detect the existence of secret data. In the current digital world, image and video are commonly used covers, widely applied in digital communication \cite{dc}, copyright protection \cite{cpr}, information certification \cite{if}, e-commerce \cite{ec}, and many other practical fields~\cite{survey,cry_book}.
% Accordingly, video steganography~\cite{video_survey} refers to the process of hiding secret information into videos and then recovering.

% Traditional steganography methods usually perform message hiding in the spatial domain~\cite{sh0,sh2,sh3,sh4,sh5} by manually adjusting the pixel values of cover image, or in the frequency domain~\cite{dft,dct,dwt} by modifying certain frequency coefficients of the transformed cover image. However, traditional methods have low hiding capacity and invisibility, easily producing artificial marking on the surface and being cracked by steganalysis methods~\cite{crack1,crack2,crack3}.

Traditional video steganography methods usually hide messages in the spatial domain or transform domain by manual design. Video steganography in the spatial domain means embedding is done directly to the pixel values of video frames. 
% ~\cite{sh0,sh2,sh3,sh4,sh5} by manually adjusting the pixel values of cover image, or in the frequency domain~\cite{dft,dct,dwt} by modifying certain frequency coefficients of the transformed cover image. 
Least significant bits (LSB)~\cite{lsb, lsb2} is the most well-known spatial-domain method,
% . In LSB, the secret data is embedded by 
replacing the $n$ least significant bits of the cover image with the most significant $n$ bits of the secret data.
% , and m
Many researchers have used LSB replacement~\cite{lsb_r} and LSB matching~\cite{lsb_m} for video steganography. The transform-domain hiding~\cite{dft,dct,dwt} is done by modifying certain frequency coefficients of the transformed frames. For instance, \cite{svideo_dct} proposed a video steganography technique by manipulating the quantized coefficients of DCT (Discrete Cosine Transformation). \cite{svideo_dwt} proposed to compare the DWT (Discrete Wavelet Transformation) coefficients of the secret image and the cover video for hiding. However, these traditional methods have low hiding capacity and invisibility, easily being cracked by steganalysis methods~\cite{crack1,crack2,crack3}.

Recently, some deep-learning methods were proposed to improve the hiding capacity and performance. Early works are presented in image steganography. Baluja~\cite{nips,tpami} proposed the first deep-learning method to hide a full-size image into another image. 
% This work was then extended in~\cite{tpami} by permuting the pixels of the secret image to enhance data security. 
Recently, \cite{isn,hinet} 
% HiNet~\cite{hinet} and RIIS~\cite{riis} 
proposed designing the steganography model as an invertible neural network (INN)~\cite{nice,realnvp} to perform image hiding and recovering with a single model. For video steganography, Khare et al.~\cite{svideo1} first utilized back propagation neural networks to improve the performance of the LSB-based scheme. \cite{svideo} is the first deep-learning method to hide a video into another video. Unfortunately, it simply aims to hide the residual across adjacent frames in a frame-by-frame manner, and it requires several separate steps to complete the video hiding and recovering. \cite{mishra2019vstegnet} utilize 3D-CNN to explore the temporal correlation in video hiding. However, it utilizes two separated 3D UNet to perform hiding and recovering, and it has high model complexity ($367.2$ million parameters). While video steganography has achieved impressive success in terms of hiding capacity to hide a full-size video, the more challenging multiple videos hiding has hardly been studied. Also, the steganography pipeline is rigid.

% Unlike existing methods, 
In this paper, we study the large-capacity and flexible video steganography, as shown in Fig.~\ref{overview}. Concretely, we propose a reversible video steganography pipeline, achieving large capacity to hide/recover multiple secret videos in/from a cover video. 
% We also introduce several weight-sharing designs to our framework, leading to attractive model complexity in multiple videos steganography. 
At the same time, our model complexity is also attractive by combining several weight-sharing designs. The flexibility of our method is twofold. First, we propose a key-controllable scheme, enabling different receivers to recover particular secret videos with specific keys. Second, we propose a scalable strategy, 
% in multiple videos hiding, 
which can hide variable numbers of secret videos into a cover video with a single model and a single training session.
% Moreover, we take video hiding and recovering as a pair of inverse problems solved by a single invertible neural network (INN). 
To summarize, this work has the following contributions:
\vspace{-4pt}
\begin{itemize}
    \item We propose a large-capacity video steganography method, which can hide/recover multiple (\textbf{up to 7}) secret videos in/from a cover video. Our hiding and recovering are fully reversible via a single INN.
    \vspace{-4pt}
    \item We propose a key-controllable scheme with which different receivers can recover particular secret videos from the same cover video via specific keys.
    \vspace{-4pt}
    \item We propose a scalable embedding module, utilizing a single model and a single training session to satisfy different requirements for the number of secret videos hidden in a cover video.
    \vspace{-4pt}
    \item Extensive experiments demonstrate that our proposed method achieves state-of-the-art performance with large hiding capacity and flexibility.
\end{itemize}

\section{Related Work}
\subsection{Video Steganography}
Steganography can date back to the 15th century, whose goal is to encode a secret message in some transport mediums and covertly communicate with a potential receiver who knows the decoding protocol to recover the secret message. Since the human visual system is less sensitive to small changes in digital media, especially digital videos. Video steganography is becoming an important research area in various data-hiding technologies~\cite{survey}.

Traditional video steganography methods usually performed hiding and recovering in the spatial domain, \textit{e.g.}, Pixel Value Differencing (PVD)~\cite{svideo_pvd1,svideo_pvd2} and Least Significant Bits (LSB)~\cite{lsb_video1,lsb_r,lsb_m}. PVD embeds the secret data in the difference value of two adjacent pixels. In~\cite{svideo_pvd1}, a PVD-based video steganography system is proposed to embed the secret data in a compressed domain of the cover medium. \cite{svideo_pvd2} utilized enhanced pixel-value differencing (EPVD) to improve the video steganography performance. LSB methods work by replacing the $n$ least significant bits of the cover data with the most significant $n$ bits of the secret information. \cite{lsb_video1} utilized LSB replacement technique to hide secret text in grayscale video frames. To enhance the security in LSB-based methods, \cite{lsb_sec} shuffled the secret data and embedded the index of correct order into the cover video. In addition to spatial-domain methods, some transformed domain methods~\cite{svideo_dct,svideo_dwt} were proposed to perform hiding by modifying certain frequency coefficients of the transformed cover video. For instance, \cite{svideo_dct} proposed a video steganography technique by manipulating the quantized coefficients of DCT transformation. \cite{svideo_dwt} proposed to compare the DWT transformation coefficients of the secret image and the cover video for hiding. Nevertheless, the above traditional methods have low hiding capacity and invisibility, easily producing artificial markings and being cracked by steganalysis methods~\cite{crack1,crack2,crack3}.

\begin{figure*}[t]
% \small
    \centering
    \includegraphics[width=0.99\linewidth]{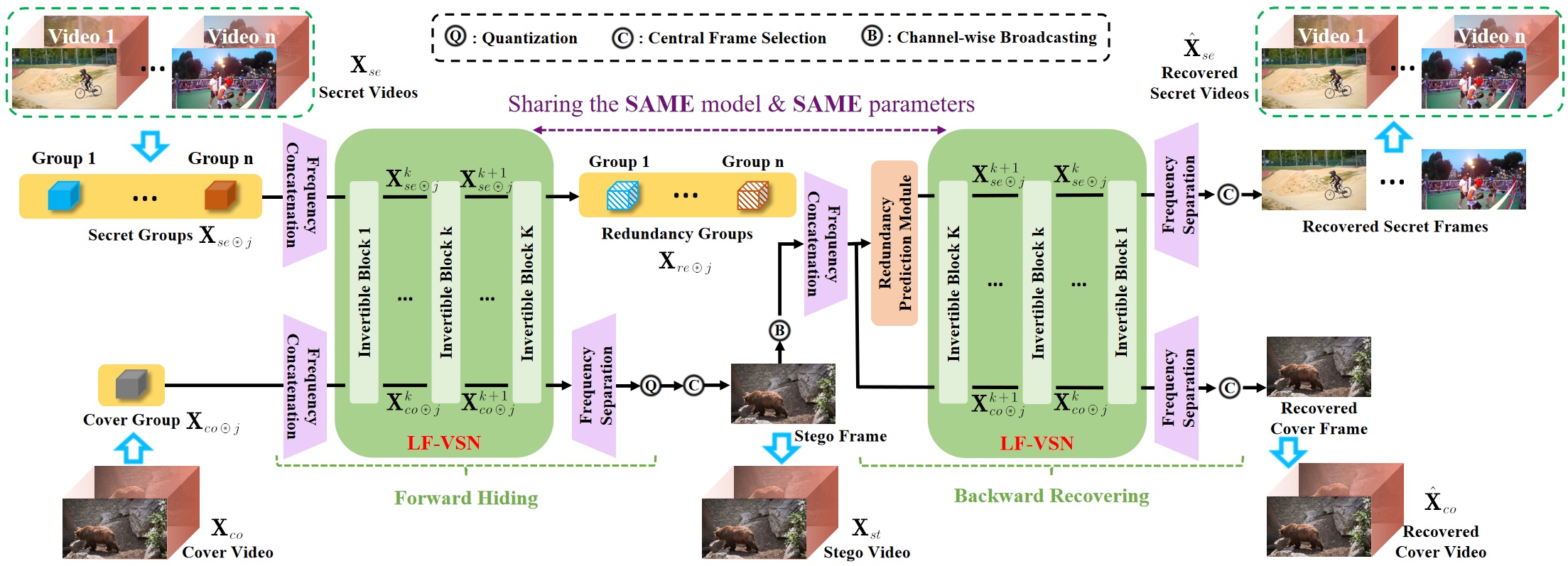}
    \vspace{-4pt}
    \caption{Network architecture of our LF-VSN. It is composed of several invertible blocks. In the forward hiding process, multiple secret videos are hidden in a cover video to generate a stego video, together with redundancy. In the backward recovering process, the stego video and predicted redundancy are fed to the reverse data flow of the same network with the same parameters to recover secret videos.
    % large-capacity and flexible video steganography network (LF-VSN). Our LF-VSN is composed of several invertible blocks to reversibly solves multiple videos hiding and recovering with a single model.
    }
    \vspace{-10pt}
    \label{network}
\end{figure*}

Motivated by the success of deep learning, some deep-learning methods were proposed. \cite{dl_s1} introduced GAN to steganography, showing that the adversarial training scheme can improve hiding security. \cite{video_w2} improve the hiding quality by utilizing two independent adversarial networks to critique the video quality and optimize for robustness. \cite{iclr} studied the lossless steganography below 3 bits per pixel (bpp) hiding.
% \cite{iclr} proposed an algorithm to take advantage of the fact that neural networks are sensitive to tiny perturbations. This work can achieved $0\%$ error reliably for hiding up to 3 bits per pixel (bpp) of secret information in images. 
\cite{video_w1} embedded the secret data in the wavelet transform coefficients of the video frames. The above methods focus more on the robustness of low-capacity hiding. One of the important applications of low-capacity steganography is watermarking~\cite{hiddn,bit_s1,water3}, in which the secret bit string represents the sign of the owner. Some deep-learning methods were proposed for large-capacity hiding. \cite{nips,tpami} first explored hiding a full-size image into another image. \cite{isn,hinet} proposed a cheaper pipeline by implementing image hiding and recovering with a single invertible neural network (INN)~\cite{nice,realnvp}. Compared with image hiding, video hiding is a more challenging task, requiring a larger hiding capacity. \cite{svideo} first studied to hide/recover a video in/from another video. However, this method simply hides the residual across adjacent frames in a frame-by-frame manner. \cite{mishra2019vstegnet} explores the temporal correlation by 3D CNN in video steganography. However, it utilizes two separate 3D UNet to perform hiding and recovering and has high model complexity ($367.2M$ model parameters). These previous works demonstrate that deep networks have great potential in video hiding, inspiring us to study the more challenging task of multiple and flexible video hiding. 

\subsection{Invertible Neural Nerwork}
Since the concept of invertible neural network (INN) was proposed in \cite{nice,realnvp}, INN has attracted more and more attention due to its pure invertible pipeline. Pioneering research on INN can be seen in image generation tasks. For instance, Glow~\cite{glow} utilized INN to construct an invertible mapping between the latent variable $\mathbf{z}$ and nature images $\mathbf{x}$. Specifically, the generative process $\textbf{x}=f_{\theta}(\textbf{z})$ given a latent variable can be specified by an INN architecture $f_{\theta}$. The direct access to the inverse mapping $\textbf{z}=f_{\theta}^{-1}(\textbf{x})$ makes inference much cheaper. Up to now, INN has been studied in several vision tasks (\textit{e.g.}, image rescaling~\cite{irn,mimo}, image restoration~\cite{inn_noise,inn_sr}, image coloring~\cite{inn_color}, and video temporal action localization~\cite{zhao2023re2tal}) and presents promising performance. 

% Due to the pure invertible pipeline requirement, the architecture of 
The architecture of INN needs to be carefully designed to guarantee the invertibility. Commonly, INN is composed of several invertible blocks, \textit{e.g.}, the coupling layer~\cite{nice}.
% is the most commonly used invertible block. 
Given the input $\mathbf{h}$, the coupling layer first splits $\mathbf{h}$ into two parts ($\mathbf{h}_1$ and $\mathbf{h}_2$) along the channel axis. Then they undergo the affine transformations with the affine parameters generated by each other:
\vspace{-4pt}
\begin{equation}
\begin{aligned}
        \hat{\mathbf{h}}_{1} &= \mathbf{h}_1\cdot \psi_{1}(\mathbf{h}_{2})+\phi_{1}(\mathbf{h}_2)\\ 
        \hat{\mathbf{h}}_{2} &= \mathbf{h}_2\cdot \psi_{2}(\hat{\mathbf{h}}_{1})+\phi_{2}(\hat{\mathbf{h}}_1),
\end{aligned}
    \label{inn_f}
\end{equation}
where $\psi(\cdot)$ and $\phi(\cdot)$ are arbitrary functions. $\hat{\mathbf{h}}_{1}$ and $\hat{\mathbf{h}}_{2}$ are the outputs of the coupling layer. Correspondingly, the inverse process is defined as:
\vspace{-4pt}
\begin{equation}
\begin{aligned}
         \mathbf{h}_{1} = \frac{\hat{\mathbf{h}}_1-\phi_{1}(\mathbf{h}_2)}{\psi_{1}(\mathbf{h}_{2})};\ \ \ \ \ \  \mathbf{h}_{2} = \frac{\hat{\mathbf{h}}_2-\phi_{2}(\hat{\mathbf{h}_1})}{\psi_{2}(\hat{\mathbf{h}}_{1})}.
\end{aligned}
    \label{inn_b}
\end{equation}

In this paper, we employ the reversible forward and backward processes of INN to perform multiple videos hiding and recovering, respectively.
% invertibility of INN to perform multiple videos hiding and recovering with a single model. 
We further improve INN to explore flexible video steganography.

\section{Methodology}
\subsection{Overview}
An overview of our LF-VSN is presented in Fig.~\ref{network}. Specifically, given $N_{s}$ secret videos $\mathbf{x}_{se} = \{ \mathbf{x}_{se}(n) \}_{n=1}^{N_{s}}$ and a cover video $\mathbf{x}_{co}$, the forward hiding is operated group-by-group through a sliding window, traversing each video from head to tail. After hiding, a stego video $\mathbf{x}_{st}$ is produced, ostensibly indistinguishable from $\mathbf{x}_{co}$ to ensure that $\mathbf{x}_{se}$ is undetectable. In the backward recovering, a channel-wise broadcasting operation ($\mathbb{R}^{3\times W\times H}\stackrel{copy}{\longrightarrow}\mathbb{R}^{3L\times W\times H}$) copies each stego frame in the channel dimension to form the reversed input. During recovering, multiple secret videos are recovered frame-by-frame in parallel. It is worth noting that the forward hiding and backward recovering share the same model architecture and parameters. 

\subsection{Steganography Input and Output Design}
At the beginning of each hiding step, a fusion module is applied to fuse frames in each group to take advantage of the inner temporal correlation. Considering that it is easy to produce texture artifacts and color distortion when hiding in the spatial dimension~\cite{crack1,hinet}, we perform the fusion by a frequency concatenation. Specifically, given the $j$-th cover group $\mathbf{X}_{co\circledast j}\in \mathbb{R}^{{L}\times 3\times W\times H}$ and secret groups $\{ \mathbf{X}_{se\circledast j}(n)\in \mathbb{R}^{{L}\times 3\times W\times H} \}_{n=1}^{N_s}$ (each contains $L$ frames), we adopt Haar discrete wavelet transform (DWT) to split each frame into four frequency bands (\textit{i.e.}, LL, HL, LH, HH). In each frame group, we concatenate the part in the same frequency band from different frames in the channel dimension and then concatenate these four bands in order of frequency magnitude, producing the final secret input $\{ \mathbf{X}_{se\circledast j}(n)\in \mathbb{R}^{12{L}_{s}\times \frac{W}{2}\times \frac{H}{2}} \}_{n=1}^{N_s}$ and cover input $\mathbf{X}_{co\circledast j}\in \mathbb{R}^{12{L}_{c}\times \frac{W}{2}\times \frac{H}{2}}$. The output of the forward hiding comprises a stego group $\mathbf{X}_{st\circledast j}$ and several redundancy groups $\{ \mathbf{X}_{re\circledast j}(n) \}_{n=1}^{N_s}$. $\mathbf{X}_{st\circledast j}$ is converted from the frequency domain to the spatial domain by a frequency separation, \textit{i.e.}, the inverse of the frequency concatenation. 
$\mathbf{X}_{re\circledast j}(n)$ represents the redundancy of the
% secret group 
$\mathbf{X}_{se\circledast j}(n)$ that does not need to be hidden and will be discarded. In our LF-VSN, we utilize the adjacent frames to cooperate with hiding the central frame. Thus, we only output the central stego frame in each hiding step. The backward recovering is similarly operated in the frequency domain and converted to the spatial domain at the output.

\subsection{Invertible Block}
\begin{figure}[t]
    \centering
    \includegraphics[width=0.9\linewidth]{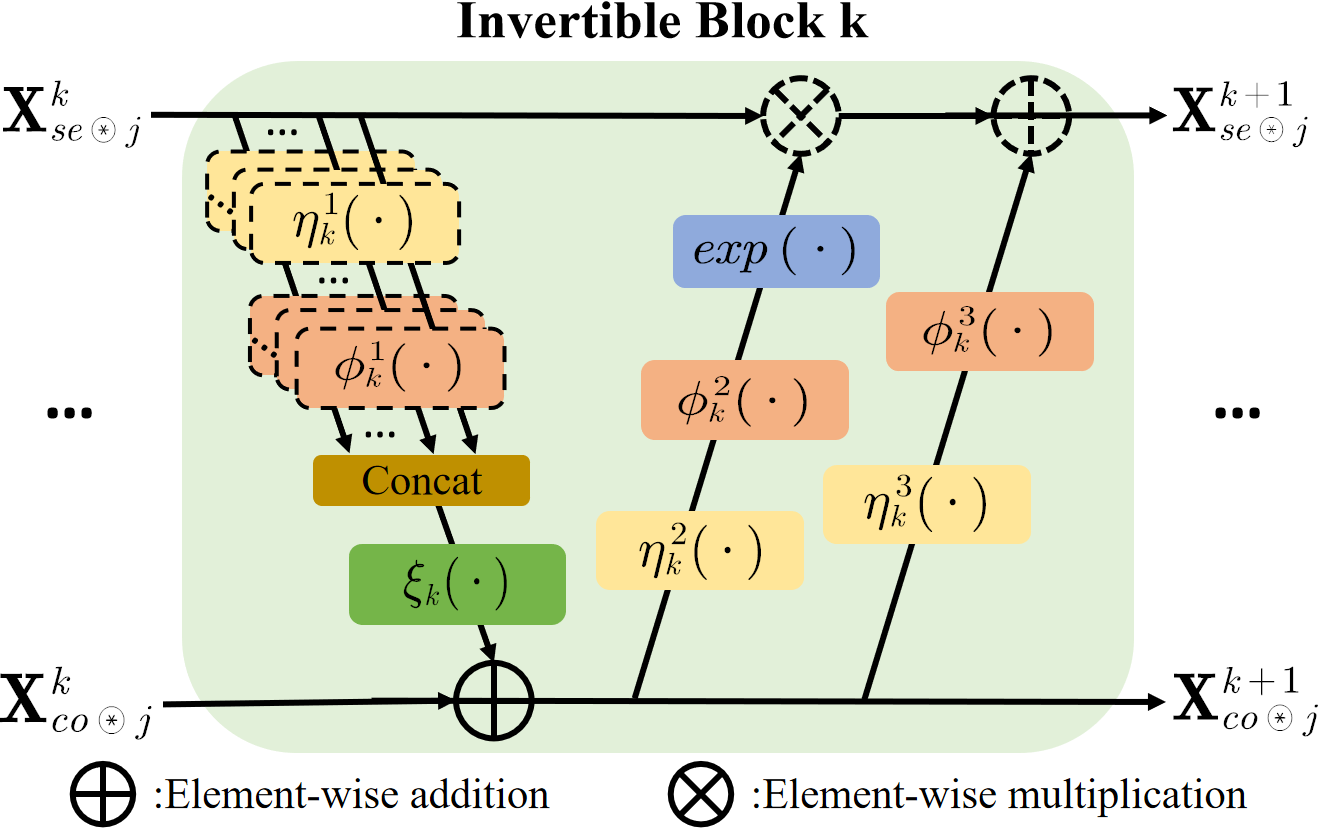}
    \vspace{-4pt}
    \caption{Illustration of the architecture of our invertible block. The dashed line refers to weight sharing.
    }
    \label{ib}
    \vspace{-10pt}
\end{figure}

As shown in Fig.~\ref{network}, our hiding and recovering have reverse information flow constructed by several invertible blocks (IBs). The architecture of IB is presented in Fig.~\ref{ib}. Concretely, in the $k$-th IB, there are two branches to process the input cover group $\mathbf{X}_{co\circledast j}^{k}$ and secret groups $\{ \mathbf{X}_{se\circledast j}^{k}(n) \}_{n=1}^{N_s}$, respectively. Several interaction pathways between these two branches construct the invertible projection. We use an additive transformation to project the cover branch and employ an enhanced affine transformation to project the secret branch. The transformation parameters are generated from each other. 
% Here we utilize several weight-sharing designs to reduce the model complexity. Specifically, 
Here we utilize weight-sharing modules ($\eta_k^{1}(\cdot)$ and $\phi_{k}^{1}(\cdot) $) to extract features from all secret groups, producing a feature set $\{ \mathbf{F}_{se}^{k}(n)\}_{n=1}^{N_s}=\{ \phi_{k}(\eta_k (\mathbf{X}_{se\circledast j}^{k}(n)))\}_{n=1}^{N_s}$. $\eta_k^{i}(\cdot)$ and $\phi_{k}^{i}(\cdot)$ ($i=1,2,3$) refer to a $3\times 3$ convolution layer and a five-layer dense block~\cite{densenet}, respectively. Then, we concatenate $\mathbf{F}_{se}^{k}$ in the channel dimension and pass through an aggregation module $\xi_{k}(\cdot)$ to generate the transformation parameters of the cover branch. Note that $\xi_{k}(\cdot)$ is optional in different cases. In our fixed hiding, $\xi_{k}(\cdot)$ is a $3\times 3$ convolution layer, and it is a scalable embedding module in our scalable hiding. The transformation parameters of the secret branch are generated from $\mathbf{X}_{co\circledast j}^{k}$ and shared among different secret groups. Thus, in the $k$-th invertible block, the bijection of the forward propagation in Eq.~\eqref{inn_f} is reformulated as: 
\begin{equation}
\small
    \begin{aligned}
        &\mathbf{X}_{co\circledast j}^{k+1} = \mathbf{X}_{co\circledast j}^{k} + \xi_{k} (||\phi_{k}^{1}(\eta_{k}^{1}(\mathbf{X}_{se\circledast j}^{k}(n)))||_{n=1}^{N_s})\\
        &\{\mathbf{X}_{se\circledast j}^{k+1}(n)\}_{n=1}^{N_s} =\\ 
        &\mathbf{X}_{se\circledast j}^{k}(n)\cdot \exp(\phi_{k}^{2}(\eta_{k}^2(\mathbf{X}_{co\circledast j}^{k+1}))) + \phi_{k}^{3}(\eta_k^3(\mathbf{X}_{co\circledast j}^{k+1})),
        % , t=[1, N_s] ,
    \end{aligned}
    \label{inn_f2}
\end{equation}
where $||\cdot ||$ refers to the channel-wise concatenation. $\exp(\cdot)$ is the Exponential function. Accordingly, the backward propagation is defined as:
\begin{equation}
\small
    \begin{aligned}
    &\{ \mathbf{X}_{se\circledast j}^{k}(n)\}_{n=1}^{N_s} =\\ 
    &(\mathbf{X}_{se\circledast j}^{k+1}(n) - \phi_{k}^{3}(\eta_k^3(\mathbf{X}_{co\circledast j}^{k+1}))) \cdot \exp(-\phi_{k}^{2}(\eta_{k}^2(\mathbf{X}_{co\circledast j}^{k+1})))\\
    &\mathbf{X}_{co\circledast j}^{k} = \mathbf{X}_{co\circledast j}^{k+1} - \xi_{k} (||\phi_{k}^{1}(\eta_{k}^{1}(\mathbf{X}_{se\circledast j}^{k}(n)))||_{n=1}^{N_s}).
    \end{aligned}
    \label{inn_f0}
\end{equation}

\begin{figure}[t]
    \centering
    \includegraphics[width=0.9\linewidth]{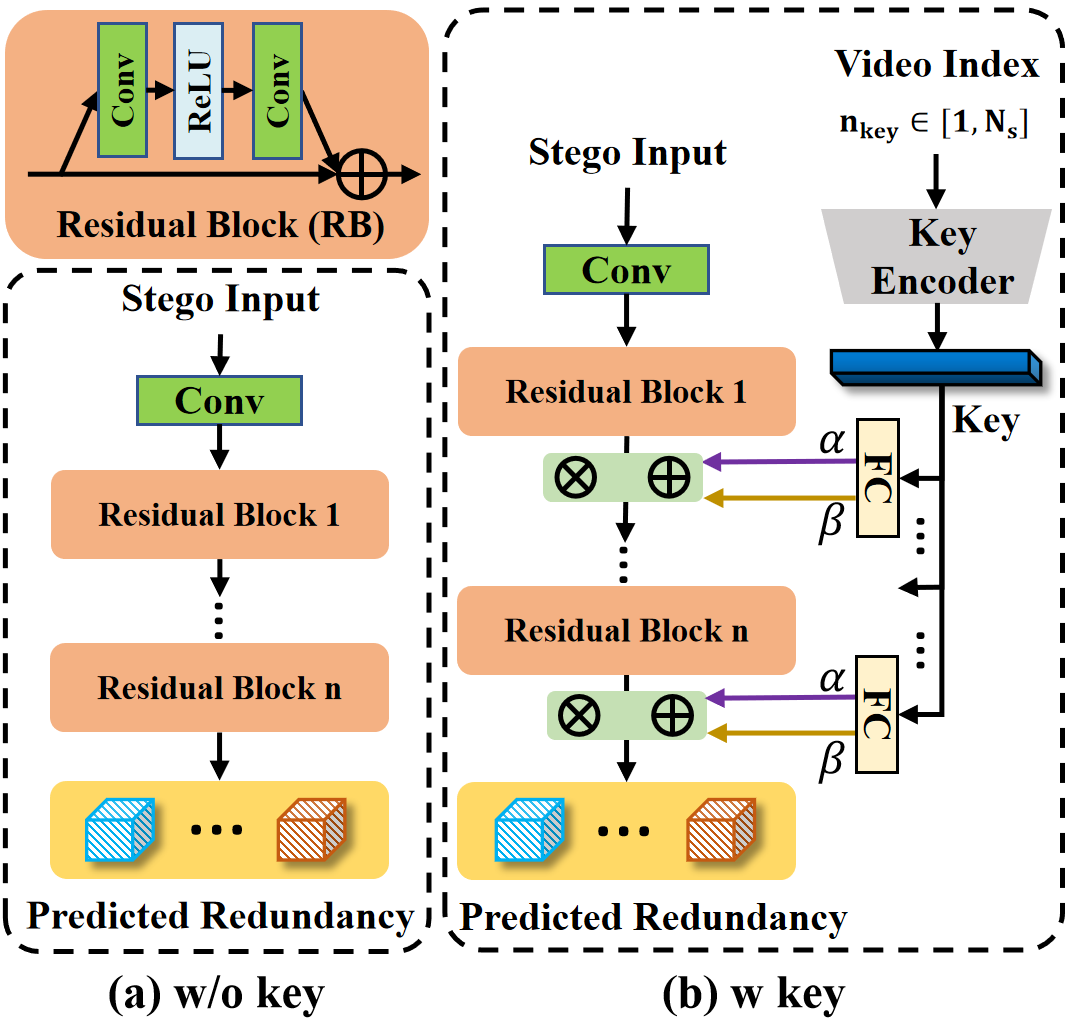}
    \vspace{-6pt}
    \caption{The architecture of our redundancy prediction module (RPM). It 
    % is composed of several residual blocks and 
    has two model settings: (a) RPM without (w/o) key controlling; (b) RPM with (w) key controlling.
    }
    \vspace{-10pt}
    \label{rpm}
\end{figure}

\subsection{Redundancy Prediction Module (RPM) \& Key-controllable Design}
As illustrated previously, we retain the stego part and discard the redundancy information in the forward hiding. Therefore, we need to prepare a suitable redundancy in the backward process to utilize the reversibility of INN to reconstruct the forward input (\textit{i.e.}, secret and cover). In different tasks, most INN-based methods~\cite{glow,irn,isn,hinet} constrain the generated redundancy information to obey the Gaussian distribution and utilize random Gaussian sampling
% sample randomly in the Gaussian distribution 
to approximate this part in the backward process. Nevertheless, such random sampling lacks data specificity and adaptivity. In our LF-VSN, we predict the redundancy information from the stego group through a redundancy prediction module (RPM), as shown in Fig.~\ref{rpm}(a). It is composed of several residual blocks (RB) without the Batch Normalization layer.

In this paper, we present a novel extension of RPM to construct key-controllable video steganography, with which we can hide multiple secret videos in a cover video and recover a secret video conditioned on a specific key. The architecture is shown in Fig.~\ref{rpm}(b). Given the index $n_{key}$ of a secret video $\mathbf{X}_{se}(n_{key})$, a specific key is generated by a key encoder, which is composed of several fully connected (FC) layers. The key is then fed into a FC layer at the end of each RB in RPM to generate a condition vector with $2C_{rpm}$ channels, which is divided into two modulation vectors $\mathbf{\alpha} , \mathbf{\beta} \in \mathbb{R}^{C_{rpm}\times 1\times 1}$ in the channel dimension. $C_{rpm}$ is the feature channel of each RB in RPM. Then we modulate the output feature $\mathbf{F}_{rpm}$ of each RB as $\mathbf{F}_{rpm}\cdot \mathbf{\alpha}+\mathbf{\beta}$. In the training process, we constrain the recovered output the same as the $n_{key}$-th secret video (\textit{i.e.}, $\mathbf{X}_{se}(n_{key})$). More details can be found in Sec.~\ref{train}.

\begin{figure}[t]
    \centering
    \includegraphics[width=0.9\linewidth]{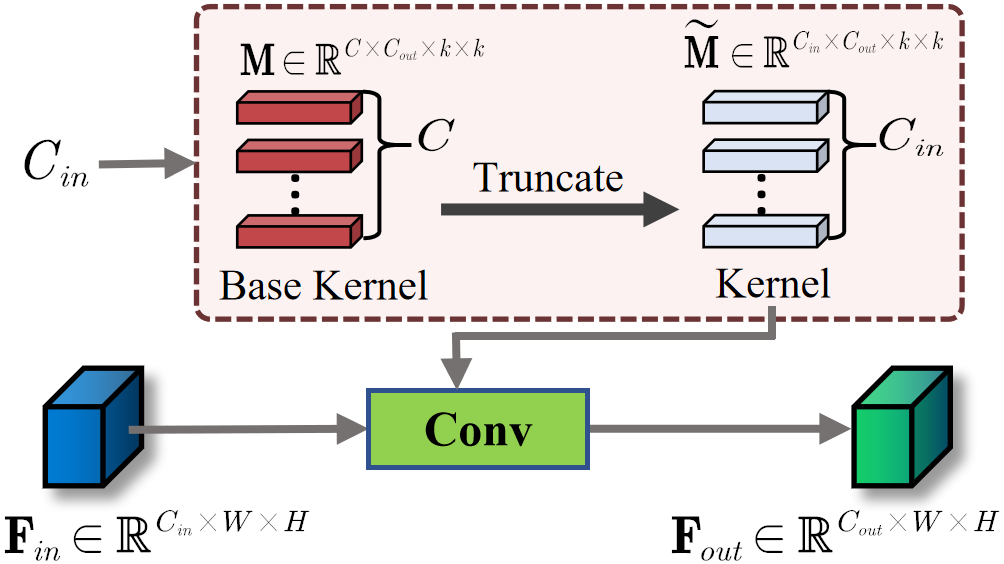}
    \caption{Illustration of our scalable embedding module. It takes the input feature map with scalable channels $C_{in}\in [1,C]$ and produces output features with fixed channels $C_{out}$.
    }
    \vspace{-10pt}
    \label{scalable}
\end{figure}

\begin{table*}[t]
\caption{Quantitative comparison (PSNR/SSIM) on Vimeo-T200. The best and second-best results are \textbf{highlighted} and \underline{underlined}. Our LF-VSN achieves the best performance in stego and secret quality with acceptable complexity. 
}
\vspace{-6pt}
\centering
\begin{tabular}{c c c c c c c c}
% \hline
\toprule
\rowcolor{color3} & Weng \textit{et al.}\cite{svideo} & Baluja\cite{tpami} & ISN\cite{isn} & HiNet\cite{hinet} & RIIS\cite{riis} & PIH\cite{pih} & LF-VSN (Ours) \\ 
\hline
\hline
Stego & 29.43/0.862 & 34.14/0.860 & 42.08/\underline{0.965} & 42.09/0.962 & \underline{43.50}/0.951 & - & \textbf{45.17}/\textbf{0.980}\\ 
\hline
Secret & 32.08/0.899 & 35.21/0.931 & 42.11/0.984 & \underline{44.44}/\underline{0.991} & 44.08/0.964 & 36.48/0.939 & \textbf{48.39}/\textbf{0.996}\\ 
\hline
Params.& 42.57M & \underline{2.65}M & 3.00M & 4.05M & 8.15M & \textbf{0.67}M & 7.40M\\
% \hline
\toprule
\end{tabular}
\label{cp_q1}
\end{table*}

\begin{figure*}[t]
\centering
\begin{minipage}[t]{0.02\linewidth}
\centering
% \vspace{3pt}
\rotatebox{90}{\normalsize{Secret video 2}}\\~\\
% \vspace{4pt}
\rotatebox{90}{\normalsize{Secret video 4}}
\end{minipage}
\begin{minipage}[t]{0.24\linewidth}
\centering
\includegraphics[width=1\columnwidth]{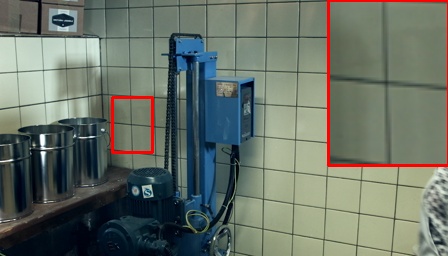}\\ 
% GroundTruth \\ \vspace{2pt}
\includegraphics[width=1\columnwidth]{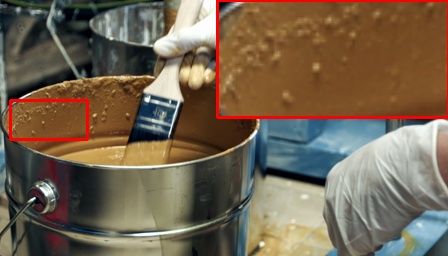}\\ GroundTruth
\end{minipage}
\begin{minipage}[t]{0.24\linewidth}
\centering
\includegraphics[width=1\columnwidth]{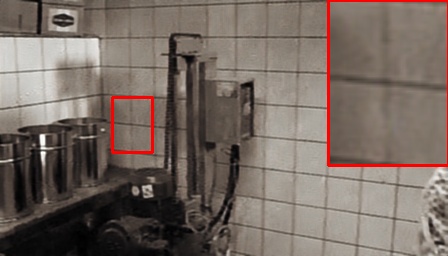}\\ 
% ISN~\cite{isn} \\ \vspace{2pt}
\includegraphics[width=1\columnwidth]{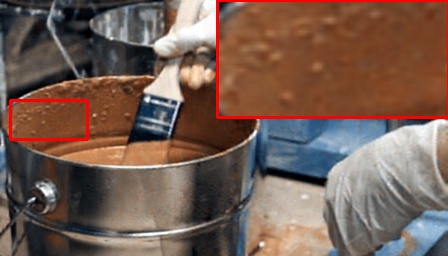}\\ ISN~\cite{isn}
\end{minipage}
\begin{minipage}[t]{0.24\linewidth}
\centering
\includegraphics[width=1\columnwidth]{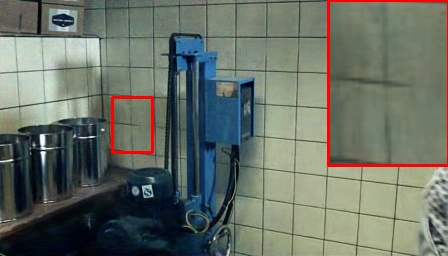}\\ 
% PIH~\cite{pih} \\ \vspace{2pt}
\includegraphics[width=1\columnwidth]{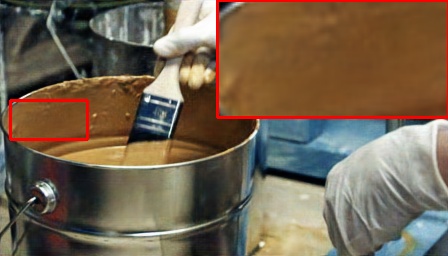}\\ PIH~\cite{pih}
\end{minipage}
\begin{minipage}[t]{0.24\linewidth}
\centering
\includegraphics[width=1\columnwidth]{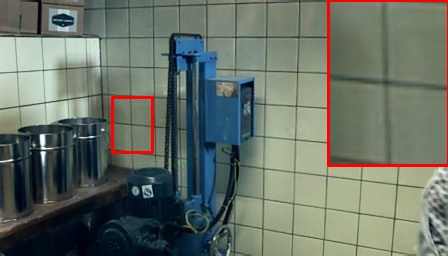}\\ 
% LF-VSN (Ours) \\ \vspace{2pt}
\includegraphics[width=1\columnwidth]{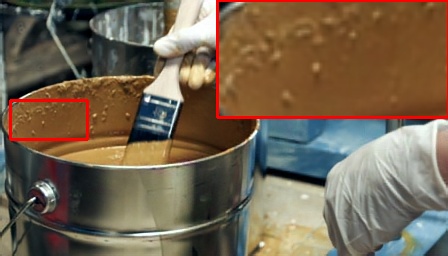}\\ LF-VSN (Ours)
\end{minipage}
\centering
\vspace{-6pt}
\caption{Visual comparison between our LF-VSN, ISN~\cite{isn}, and PIH~\cite{pih} in $4$ videos Steganography. We present the secret reconstruction results of video $2$ and video $4$. Our LF-VSN produces better result with intact color and details.
}
\vspace{-6pt}
% \vspace{-2pt}
\label{vs-vs-mul2} 
\end{figure*}

\begin{figure}[t]
\centering
\begin{minipage}[t]{0.242\linewidth}
\centering
\includegraphics[width=1\columnwidth,height=0.64\columnwidth]{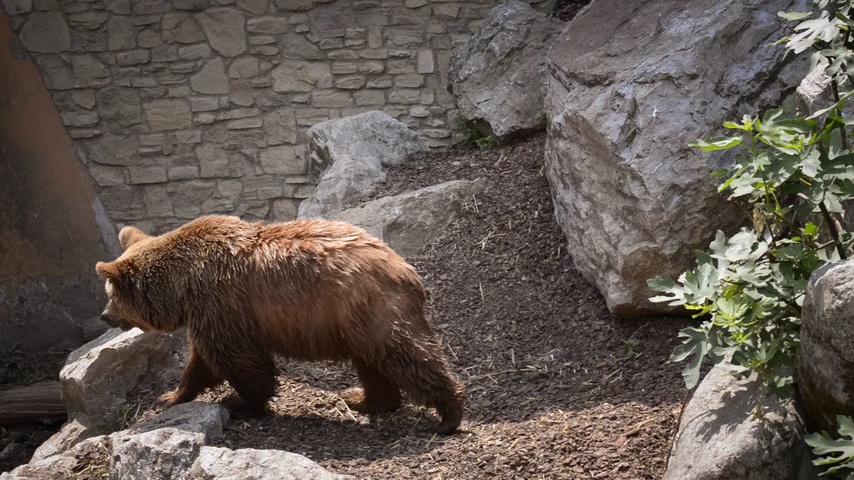}\\ $\mathbf{X}_{co}$\\ %\vspace{2pt}
\includegraphics[width=1\columnwidth,height=0.64\columnwidth]{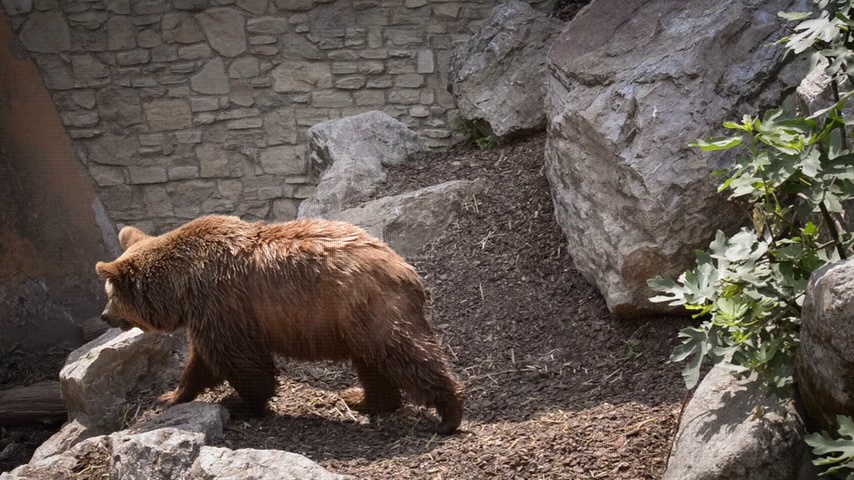}\\ $\mathbf{X}_{st}$
\end{minipage}
\begin{minipage}[t]{0.242\linewidth}
\centering
\includegraphics[width=1\columnwidth,height=0.64\columnwidth]{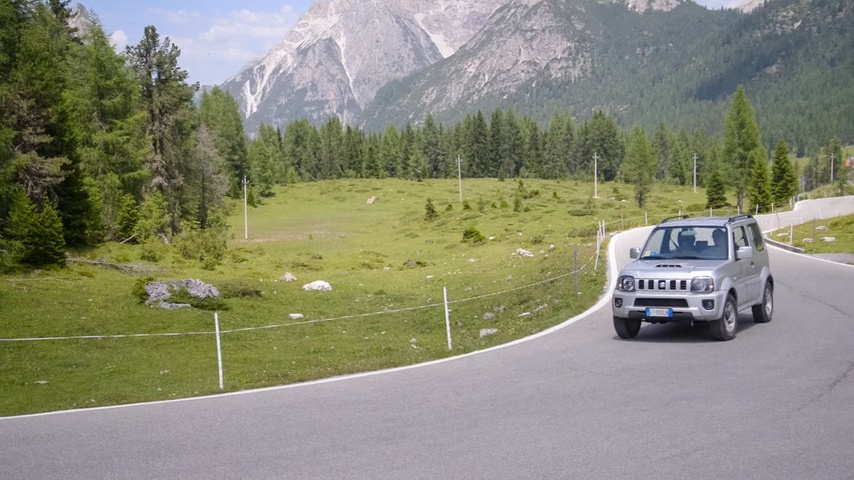}\\ $\mathbf{X}_{se}(3)$ \\ %\vspace{2pt}
\includegraphics[width=1\columnwidth,height=0.64\columnwidth]{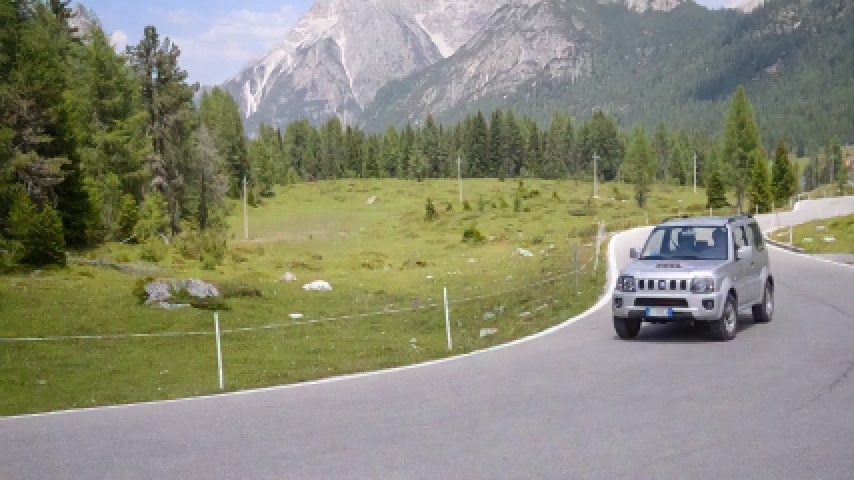}\\ $\hat{\mathbf{X}}_{se}(3)$
\end{minipage}
\begin{minipage}[t]{0.242\linewidth}
\centering
\includegraphics[width=1\columnwidth,height=0.64\columnwidth]{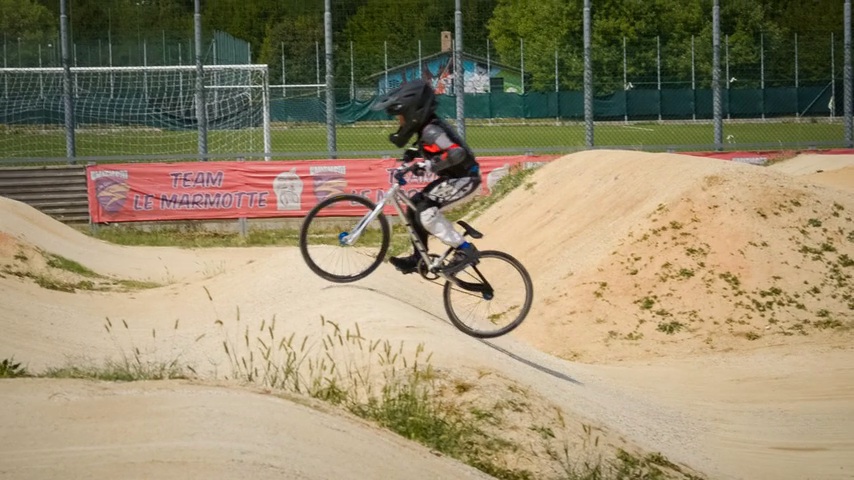}\\ $\mathbf{X}_{se}(5)$ \\ %\vspace{2pt}
\includegraphics[width=1\columnwidth,height=0.64\columnwidth]{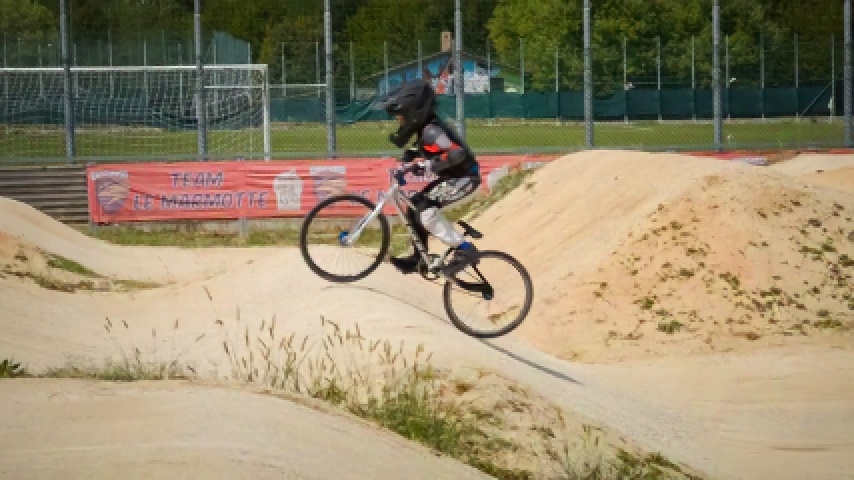}\\ $\hat{\mathbf{X}}_{se}(5)$
\end{minipage}
\begin{minipage}[t]{0.242\linewidth}
\centering
\includegraphics[width=1\columnwidth,height=0.64\columnwidth]{imgs/ours_7vi/19_6_LRGT.jpg}\\ $\mathbf{X}_{se}(7)$ \\ %\vspace{2pt}
\includegraphics[width=1\columnwidth,height=0.64\columnwidth]{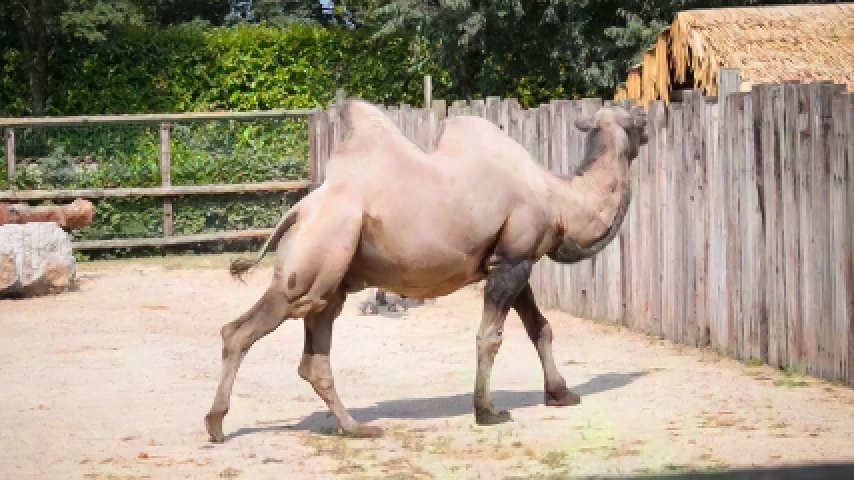}\\ $\hat{\mathbf{X}}_{se}(7)$
\end{minipage}
\centering
\caption{Visualization of our LF-VSN in $7$ videos steganography, showing promising performance in such an extreme case.
}
\vspace{-6pt}
\label{vs-vs-mul} 
\end{figure}

\subsection{Scalable Embedding Module}
The scalable design is used to handle the case where there are different requirements for the number of secret videos hidden in a cover video. It is succinctly designed on the feature aggregation part $\xi_{k}(\cdot)$ in each IB, as shown in Fig.~\ref{ib}. The illustration of our scalable embedding module is presented in Fig.~\ref{scalable}. It can be regarded as a special convolution layer, whose dimension of the convolution kernel can be changed according to the input. All convolution kernels $\widetilde{\mathbf{M}}$ with different dimensions are parameter-shared from the same base kernel $\mathbf{M}$. Technically, given the input feature $\mathbf{F}_{in}\in \mathbb{R}^{C_{in}\times W\times H}$, we truncate a convolution kernel $\widetilde{\mathbf{M}}\in \mathbb{R}^{C_{in}\times C_{out}\times k\times k}$ from $\mathbf{M}\in \mathbb{R}^{C\times C_{out}\times k\times k}$ to match the input dimension and then perform convolution: $\mathbf{F}_{out}=\widetilde{\mathbf{M}}*\mathbf{F}_{in}$. In this way, the training of $\mathbf{M}$ is completed through the training of all sub-kernels $\widetilde{\mathbf{M}}$.

\subsection{Loss Function}
\label{train}
In our LF-VSN, the loss function is used to constrain two parts, \textit{i.e.}, forward hiding and backward recovering. The forward hiding is to hide multiple secret videos in the cover video. The generated stego video $\mathbf{X}_{st}$ should be undetectable to secret videos and as similar as possible to the cover video.  Therefore, we constrain $\mathbf{X}_{st}$ to be the same as the cover video $\mathbf{X}_{co}$:
\begin{equation}
    \mathcal{L}_{f} = ||\mathbf{X}_{st\circledast j}[I_c]-\mathbf{X}_{co\circledast j}[I_c]||_2^2,
\end{equation}
where $||\cdot||_2^2$ donates the $\ell_2$ norm. $I_c$ is the index of the central frame in each group. In the backward recovering, there are two patterns: with and without key controlling. In both patterns, we aim to recover the secret information from the cover video. The difference stands between recovering a specific secret video and all secret videos. In the pattern without key controlling, the loss function is defined as:
\begin{equation}
\begin{split}
    \mathcal{L}_{b} = \sum_{n=1}^{N_s}||\hat{\mathbf{X}}_{se\circledast j}(n)[I_c]-\mathbf{X}_{se\circledast j}(n)[I_c]||_2^2 +\\ ||\hat{\mathbf{X}}_{co\circledast j}[I_c]-\mathbf{X}_{co\circledast j}[I_c]||_2^2,
\end{split}
\end{equation}
where $\hat{\mathbf{X}}_{se}$ and $\hat{\mathbf{X}}_{co}$ represent the recovered secret and cover videos. In the pattern with key controlling, the loss function is defined to guarantee that the key generated from the video index $n_{key}$ can only recover the $n_{key}$-th secret video. Thus, the loss function is reformulated as:
\begin{equation}
\begin{split}
    \mathcal{L}_{b} = ||\frac{1}{N_s}\sum_{n=1}^{N_s}\hat{\mathbf{X}}_{se\circledast j}(n)[I_c]-\mathbf{X}_{se\circledast j}(n_{key})[I_c]||_2^2 +\\ ||\hat{\mathbf{X}}_{co\circledast j}[I_c]-\mathbf{X}_{co\circledast j}[I_c]||_2^2.
\end{split}
\end{equation}

We optimize our LF-VSN by minimizing the forward loss function $\mathcal{L}_{f}$ and backward loss function $\mathcal{L}_{b}$ as:
\begin{equation}
    \mathcal{L} = \mathcal{L}_{f} + \lambda \mathcal{L}_{b},
\end{equation}
where $\lambda$ is a hyper-parameter to make a trade-off between forward hiding and backward recovering. We set $\lambda =4$ to balance these two loss portions.

\section{Experiment}
\subsection{Implementation Details}
In this work, we adopt the training set of Vimeo-90K~\cite{vim90k} to train our LF-VSN. Each sequence has a fixed spatial resolution of $448 \times 256$. During training, we randomly crop training videos to $144\times 144$ with random horizontal and vertical flipping to make a data augmentation. We use Adam optimizer~\cite{adam}, with $\beta_1 = 0.9$, $\beta_2 = 0.5$. We set the batch size as $16$. The weight decay factor is set as $1\times 10^{-12}$. We use an initial learning rate of $1\times 10^{-4}$, which will decrease by half for every $30K$ iterations. The number of total iterations is set as $250K$. The training process can be completed on one NVIDIA Tesla V100 GPU within 3 days. For testing, we select $200$ sequences from the testing set of Vimeo-90K, denoted as Vimeo-T200 in this paper.

\subsection{Comparison Against Other Methods}
% To present the performance of our method, 
Here we compare our LF-VSN with other methods on single video steganography and challenging multiple videos steganography. The evaluation includes the stego quality in forward hiding and the secret quality in backward recovering. For single video steganography, we compare our LF-VSN with some well-known methods~\cite{svideo,tpami} and recent proposed methods~\cite{isn,hinet,riis,pih}. 
% The results are presented in Tab.~\ref{cp_q1}. 
Note that PIH~\cite{pih} highlighted the need to quantize the stego image from the floating-point format of $32\times 3$ to $8\times 3$ bits per pixel. But PIH just added the quantization to the compared methods without retraining. Here we retrain HiNet~\cite{hinet} with quantization to make a more fair comparison. 
% to make a more fair comparison, we retrain 
% ISN~\cite{isn} and HiNet~\cite{hinet}
% the non-quantization method~\cite{hinet} 
Thus, its performance may be slightly higher than that reported in PIH. ISN~\cite{isn}, RIIS~\cite{riis} and PIH were originally designed with quantization, which can be directly compared. Tab.~\ref{cp_q1} presents that our method achieves the best performance on stego and secret while maintaining acceptable complexity. 
% More discussions can be found in the supplementary. 

% As mentioned above, our proposed LF-VSN has a large capacity, which can hide/recover multiple videos in/from one video. 
For multiple videos steganography, ISN~\cite{isn} and PIH~\cite{pih} studied how to hide multiple secret images in a cover image, which can be
% regarded as 
competitive counterparts of our LF-VSN. ISN can hide up to $5$ secret images into 1 cover image, and PIH can hide $4$ secret images. 
% We present the performance of multiple-video steganography of our LF-VSN, ISN, and PIH in 
The comparison in Tab.~\ref{vs-mul} shows the better
% and more stable 
performance of our LF-VSN. Even in the $7$ videos hiding, our method still has promising stego and secret quality ($>35dB$). We present the visual comparison of different methods
% our LF-VSN, ISN and PIH 
in $4$ videos steganography in Fig.~\ref{vs-vs-mul2}. Obviously, ISN has color distortion, and PIH has a loss of details. By contrast, our LF-VSN can recover high-fidelity results. We also present the secret and stego quality of our LF-VSN in $7$ videos hiding in Fig.~\ref{vs-vs-mul}. These videos are from DAVIS~\cite{davis} dataset. One can see that our LF-VSN has promising performance in such an extreme case.

\begin{table}[t]
\caption{Multiple videos steganography comparison (PSNR) of our LF-VSN, ISN~\cite{isn}, and PIH~\cite{pih} on Vimeo-T200 test set. Our LF-VSN can hide/recover 7 videos with promising performance.}
\vspace{-6pt}
\centering
\footnotesize
\begin{tabular}{c |>{\columncolor{color3}}c| c c c c c c}
% \hline
\toprule
& Videos & 2 & 3 & 4 & 5 & 6 & 7\\ 
% \hline
\toprule
\multirow{2}*{\rotatebox{90}{\cellcolor{color3}\footnotesize{ 
 N}}}& Stego & 37.60 & 36.41 & 32.56 & 31.46 & - & -\\ 
% \cline{2-8}
\rotatebox{90}{\cellcolor{color3}\footnotesize{ IS}}
& Secret & 41.47 & 38.76 & 33.42 & 33.39 & - & -\\ 
\toprule
\multirow{2}*{\rotatebox{90}{\cellcolor{color3}\footnotesize{PIH}}}& Stego & - & - & - & - & - & - \\ 
\rotatebox{90}{\cellcolor{color3}\footnotesize{ P}}
& Secret & 35.95 & 34.96 & 34.20 & - & - & -\\ 
% \cline{2-8}
\toprule
\multirow{2}*{\rotatebox{90}{\cellcolor{color3}\footnotesize{Ours}}}& Stego & \textbf{40.97} & \textbf{38.55} & \textbf{37.55} & \textbf{36.57} & \textbf{35.68} & \textbf{35.01}\\ 
% \cline{2-8}
\rotatebox{90}{\cellcolor{color3}\footnotesize{ O}}
& Secret & \textbf{44.24} & \textbf{42.27} & \textbf{40.21} & \textbf{38.88} & \textbf{36.94} & \textbf{35.71}\\ 
% \hline
\toprule
\end{tabular}
\vspace{-10pt}
\label{vs-mul}
\end{table}

\begin{figure}[t]
\centering
\begin{minipage}[t]{0.32\linewidth}
\centering
\includegraphics[width=1\columnwidth]{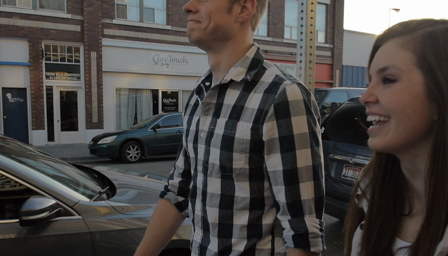}\\ Secret video 2 \\
\includegraphics[width=1\columnwidth]{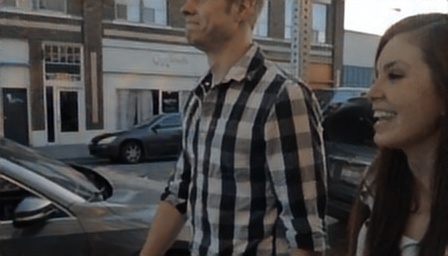}\\ Key 2 recovered\\
\includegraphics[width=1\columnwidth]{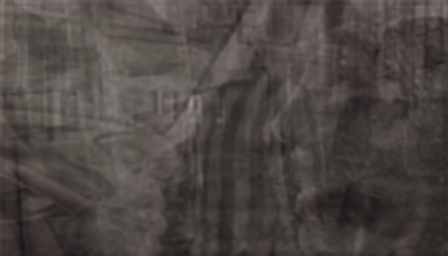}\\ Key 2\textcolor{red}{$^*$} recovered
\end{minipage}
\begin{minipage}[t]{0.32\linewidth}
\centering
\includegraphics[width=1\columnwidth]{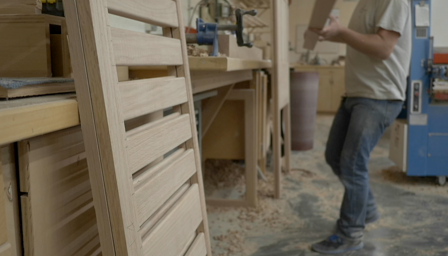}\\ Secret video 4 \\
\includegraphics[width=1\columnwidth]{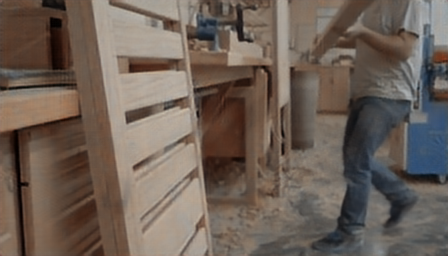}\\ Key 4 recovered \\
\includegraphics[width=1\columnwidth]{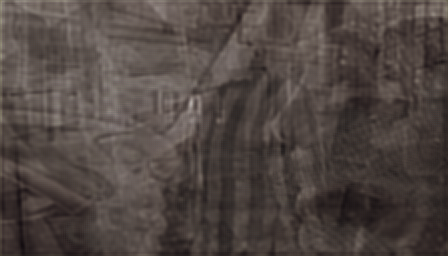}\\ Key 4\textcolor{red}{$^*$} recovered
\end{minipage}
\begin{minipage}[t]{0.32\linewidth}
\centering
\includegraphics[width=1\columnwidth]{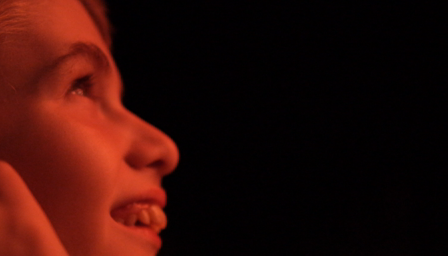}\\ Secret video 6 \\
\includegraphics[width=1\columnwidth]{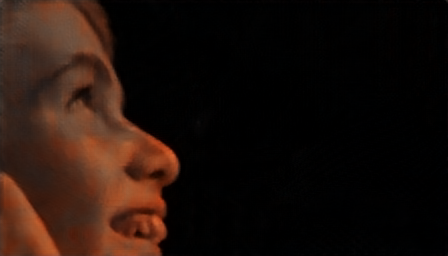}\\ Key 6 recovered\\
\includegraphics[width=1\columnwidth]{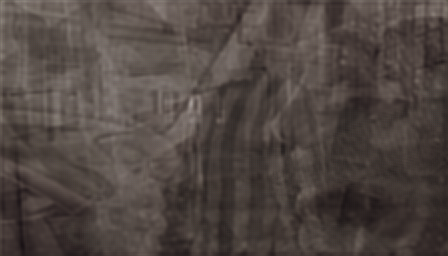}\\ Key 6\textcolor{red}{$^*$} recovered
\end{minipage}
\centering
\vspace{-6pt}
\caption{Visualization of our key-controllable scheme in $6$ videos steganography. In the second and third rows, we use the correct and wrong (\textcolor{red}{*}) keys of $2$, $4$, $6$ to recover secret videos, respectively. 
}
\vspace{-15pt}
\label{con-mul} 
\end{figure}

\subsection{Key-controllable Video Steganography}
Hiding multiple secret videos in a cover video is challenging; doing so for different receivers is even more difficult. In this paper, we present a key-controllable scheme in multiple videos steganography. It enables different receivers to recover particular secret videos through specific keys. The comparison in Tab.~\ref{vs-mul} presents that our controllable scheme still has a large hiding capacity (up to $6$ videos) with attractive performance ($>30dB$). The visualization of recovering quality is presented in the second row of Fig.~\ref{con-mul}, showing the high-quality and key-controllable results of our LF-VSN in multiple videos steganography.  

We also study the security of our controlling scheme, \textit{i.e.}, the key is sensitive and model-specific.
% generated from another set of model parameters (\textit{i.e.}, decoding protocol) can not recover the secret information. 
Here we take two sets of parameters, producing from the 250K and 240K iterations in the same training process. We use the key 
% (\textcolor{red}{$*$}) from 
produced by one model (\textcolor{red}{*}) to recover the secret video hidden by another. The result in the third row of Fig.~\ref{con-mul} presents that the wrong key has no controlling and recovering ability. Thus, our key-controllable scheme not only has the controlling function but also enhances data security.

\begin{table}[t]
\caption{Performance comparison between controllable (C) and non-controllable (NC) video steganography of our LF-VSN.}
\vspace{-6pt}
\centering
\small
% \footnotesize
\begin{tabular}{>{\columncolor{color3}}c| c c c}
% \hline
\toprule
Num. videos & 2 & 4 & 6\\ 
\hline
Stego (NC/C) & 40.97/38.67 & 37.55/34.41 & 35.68/30.48\\ 
\hline
Secret (NC/C) & 44.24/41.04 & 40.21/37.15 & 36.94/31.95\\ 
% \hline
\toprule
\end{tabular}
\label{vs-mul}
\vspace{-10pt}
\end{table}

\begin{figure}[t]
    \centering
    \begin{minipage}[t]{0.495\linewidth}
    \centering
    \includegraphics[width=1\linewidth]{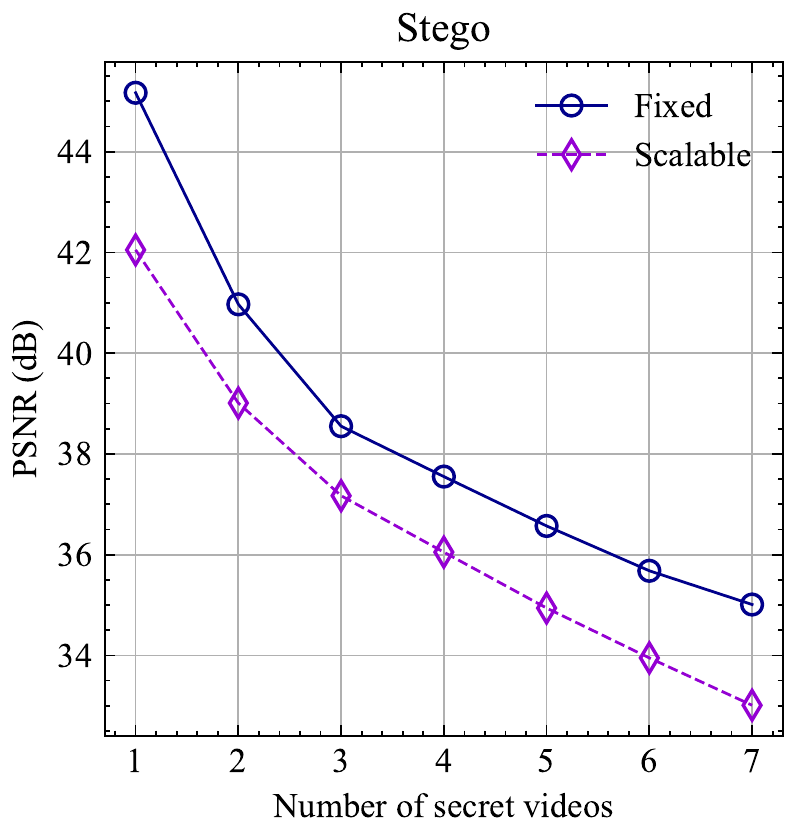}
    \end{minipage}
        \begin{minipage}[t]{0.485\linewidth}
    \centering
    \includegraphics[width=1\linewidth]{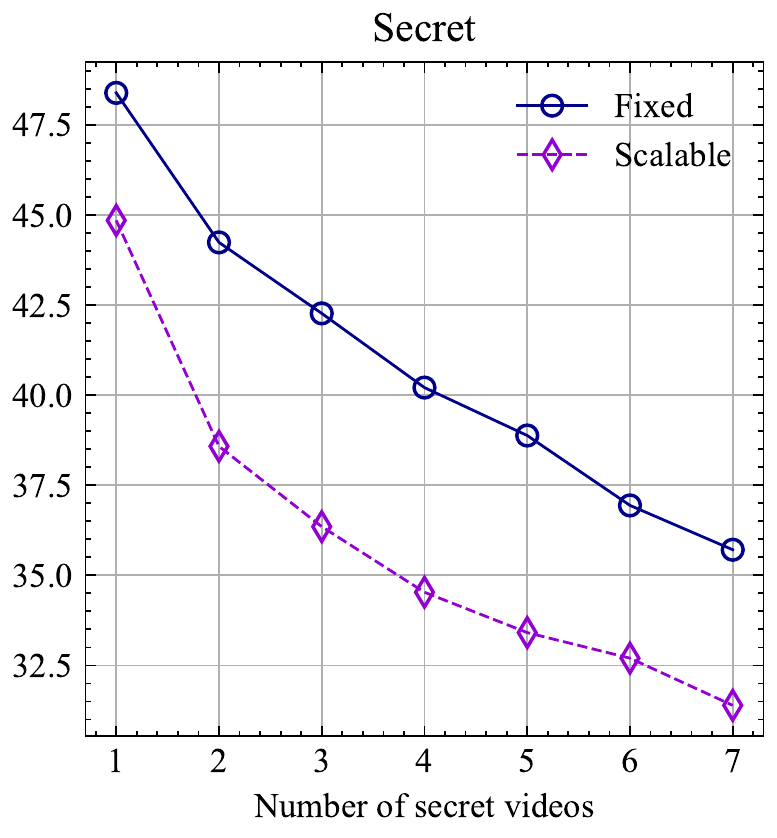}
    \end{minipage}
    \vspace{-6pt}
    \caption{
    % Illustration of the design of the scalable embedding module.
    Performance comparison between our scalable and fixed design in multiple videos steganography. 
    % It takes the input with scalable channels and outputs features with fixed channels.
    }
    \vspace{-10pt}
    \label{Tb_scalable}
\end{figure}

\begin{table*}[t]
\caption{The ablation study of different components in our LF-VSN. It includes the sliding window size, number of invertible blocks (IB), frequency concatenation (FreqCat), and  redundancy prediction module (RPM).}
\vspace{-10pt}
\centering
% \small
\footnotesize
\begin{tabular}{>{\columncolor{color3}} c|| c c c | c c c | c c c ||c c c|| c|| c}
% \hline
\toprule
Num. videos & \multicolumn{3}{c|}{2} & \multicolumn{3}{c|}{4} & \multicolumn{3}{c||}{6} & \multicolumn{3}{c||}{3} & 3 & 3\\ 
\hline
Window size & 1 & 3 & 5 & 1 & 3 & 5 & 1 & 3 & 5 & 3 & 3 (ours) & 3 & 3 & 3\\
% \hline
Num. IB & 16 & 16 & 16 & 16 & 16 & 16 & 16 & 16 & 16 & 12 & 16 & 20 & 16 & 16\\
% \hline
FreqCat & $\checkmark$ & $\checkmark$ & $\checkmark$ & $\checkmark$ & $\checkmark$ & $\checkmark$ & $\checkmark$ & $\checkmark$ & $\checkmark$ & $\checkmark$ & $\checkmark$ & $\checkmark$ & $\checkmark$ & $\times$\\
% \hline
RPM & $\checkmark$ & $\checkmark$ & $\checkmark$ & $\checkmark$ & $\checkmark$ & $\checkmark$ & $\checkmark$ & $\checkmark$ & $\checkmark$& $\checkmark$ & $\checkmark$ & $\checkmark$ & $\times$ & $\checkmark$\\
\hline
\hline
Stego & 39.64 & 40.97 & 41.08 & 36.41 & 37.55 & 37.86 & 34.47 & 35.46 & 35.96 & 38.03 & 38.55 & 38.91 & 38.28 & 36.85\\ 
% \hline
Secret & 42.97 & 44.24 & 44.43 & 37.67 & 40.21 & 40.42 & 35.11 & 36.83 & 39.97 & 41.99 & 42.27 & 42.40 & 41.69 & 40.36\\ 
% \hline
\toprule
\end{tabular}
\label{tb-ab}
\vspace{-10pt}
\end{table*}

\subsection{Scalable Video Steganography}
In this paper, we present a scalable scheme in multiple videos steganography. It can hide a variable number of secret videos into a cover video with a single model. We evaluate the performance of our scalable design and compare it with the fixed version in Tab.~\ref{Tb_scalable}. Obviously, our method has an attractive performance ($>31dB$) in hiding a variable number (up to 7) of secret videos into a cover video by a single model. The performance degradation compared to fixed version is acceptable.
% , and the performance is slightly lower than the fixed version. 
With this design, a single model can satisfy multiple steganography demands.

\begin{figure}[t]
    \centering
    \includegraphics[width=0.85\linewidth]{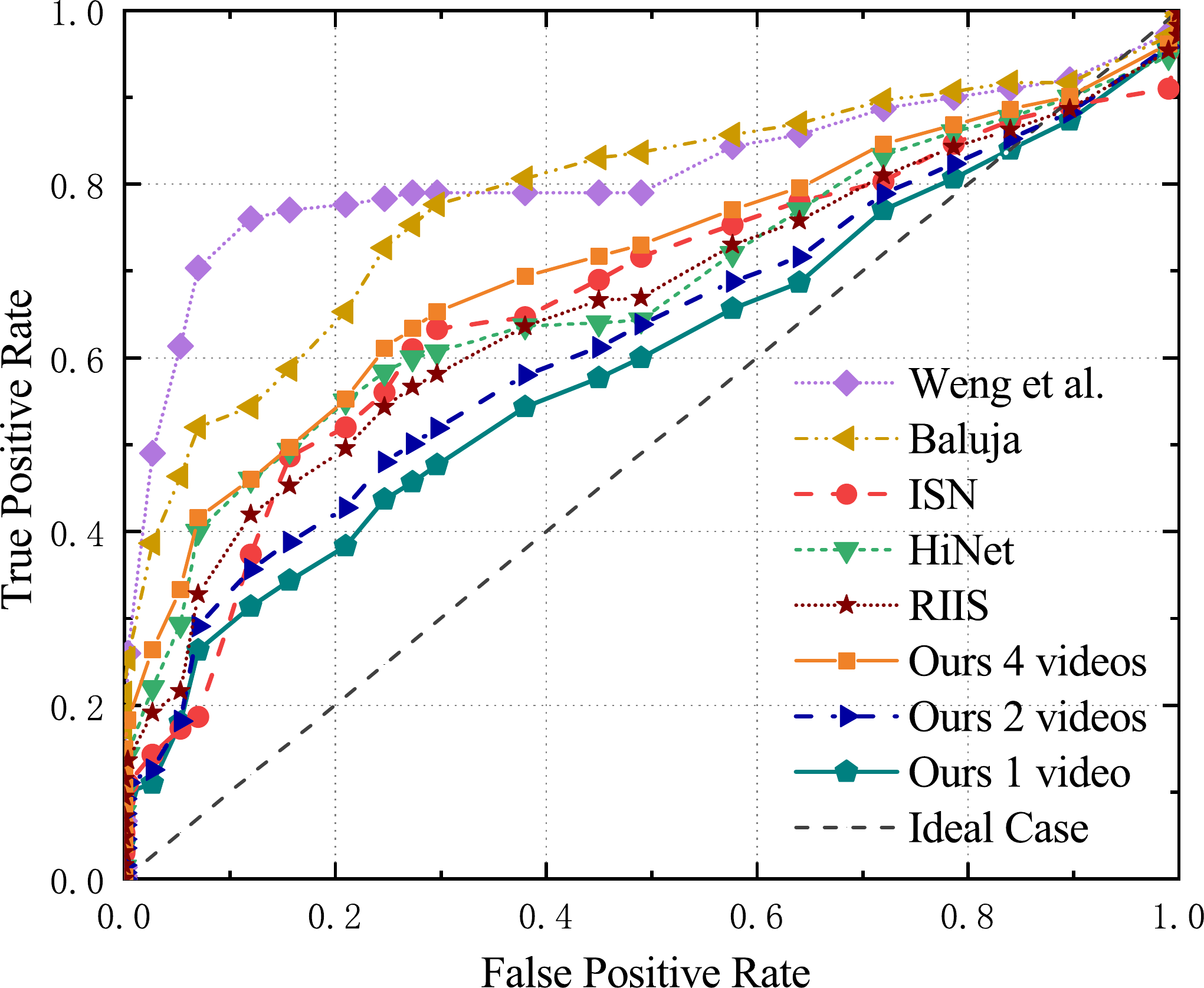}
    \vspace{-6pt}
    \caption{
    Statistics-based steganalysis by StegExpose \cite{boehm2014stegexpose}. The closer the detection accuracy is to $50\%$, the higher the security is.
    }
    \vspace{-10pt}
    \label{tb:expose}
\end{figure}

\subsection{Steganographic Analysis}
\label{Steganographic}
The data security is one of the most important concerns in steganography. In this section, we evaluate the anti-steganalysis ability of different methods, which stands for the possibility of detecting stego frames from nature frames by steganalysis tools. We utilize the StegExpose~\cite{boehm2014stegexpose} to attack different steganography methods. The detection set is built by mixing stego and cover with equal proportions. We vary the detection thresholds in a wide range in StegExpose and draw the receiver operating characteristic (ROC) curve in Fig.~\ref{tb:expose}. Note that the ideal case represents that the detector has a $50\%$ probability of detecting stego from an equally mixed cover and stego, the same as random sampling. Therefore, the closer the curve is to the ideal case, the higher the security is. Obviously, the stego frames generated by our LF-VSN are harder to be detected than other methods.
% , and the curve is close to the ideal case. 
Even in the multiple videos (\textit{e.g.,} 2 and 4 videos) hiding, our method can still achieve attractive performance, demonstrating the higher data security of our LF-VSN.

\subsection{Ablation Study}
In this subsection, we present the ablation study in Tab.~\ref{tb-ab} to investigate the effect of different components in our LF-VSN. The experiments are conducted on Vimeo-T200.

\noindent \textbf{Sliding window size.} In this paper, we utilize the temporal correlation within each frame group to improve the video steganography performance. To demonstrate the effectiveness, we evaluate the performance of our LF-VSN with the window size $L=\{ 1,3,5\} $ in 2, 4, and 6 videos steganography. The results in Tab.~\ref{tb-ab} present that the temporal correlation has obvious performance gains to the multiple videos steganography. Considering the model complexity, we set the sliding window size as $3$ in our LF-VSN.

\noindent \textbf{Number of invertible blocks (IB).} As mentioned above, our LF-VSN is composed of several IBs. 
% In this part, we study the effectiveness of IB. AS shown in Tab.~\ref{tb-ab}, 
To investigate the effectiveness of IB, we evaluate the performance of our LF-VSN with the number of IB being 12, 16, and 20. The results in Tab.~\ref{tb-ab} present that the performance increases with the number of IB.
% the positive effectiveness of IB. 
To make a trade-off between performance and complexity, we utilize $16$ IBs in our LF-VSN.

\noindent \textbf{Frequency concatenation (FreqCat).} In our LF-VSN, we use the DWT transform to merge each input group in the frequency domain. To demonstrate the effectiveness, we replace this operation with direct channel-wise concatenation. Tab.~\ref{tb-ab} presents that there are $1.7dB$ and $1.91dB$ gains of FreqCat on stego and secret quality in 3 videos steganography. The possible reason is that DWT transform can separate the low-frequency and high-frequency sub-bands, making it more effective for information fusion and hiding.

\noindent \textbf{Redundancy prediction module (RPM).} In our LF-VSN, we employ RPM to predict the redundancy in the backward process instead of randomly sampling.
% from Gaussian distribution. 
To demonstrate the effectiveness of RPM, we replace this module with a random Gaussian sampling.
% in Tab.~\ref{tb-ab}. 
The result in Tab.~\ref{tb-ab} shows that RPM can be used not only to design key-controllable steganography, but also to improve performance. 

\section{Conclusion}
In this paper, we propose a large-capacity and flexible video steganography network (LF-VSN). The novelty of our method is twofold. First, our LF-VSN has a large hiding capacity, with which we can hide \textbf{$7$ secret videos} into a cover video and then recover them well ($>35dB$). Second, we explore the flexibility in multiple videos steganography by proposing a key-controllable scheme and a scalable design. Specifically, our key-controllable scheme can enable different receivers to recover particular secret videos through specific keys. Also, the key controlling is sensitive and model-specific, which can enhance data security. Our scalable design further improves the flexibility to hide a variable number of secret videos into a cover video with a single model. Extensive experiments demonstrate that our proposed LF-VSN has state-of-the-art performance with high security, large hiding capacity, and flexibility. 

%%%%%%%%% REFERENCES
{\small
\bibliographystyle{ieee_fullname}
\bibliography{egbib}
}

\end{document}